\documentclass{article}
\usepackage{authblk}
\usepackage[showframe=False]{geometry}
\newgeometry{top=3cm, bottom=2.5cm, left=2.5cm, right=2.5cm}
\usepackage{fancyhdr}
\usepackage{sectsty}
\sectionfont{\fontsize{12}{15}\selectfont}

\usepackage{microtype}
\usepackage{graphicx}
\usepackage{booktabs} 
\usepackage[utf8]{inputenc}
\usepackage[english]{babel}
\usepackage{makecell}
\usepackage{tabularx}
\usepackage{wrapfig}
\usepackage{amsthm}
\theoremstyle{plain}
\theoremstyle{definition}
\usepackage{scalerel}
\usepackage{amsmath}
\usepackage{amssymb}

\usepackage{bm}
\usepackage{hyperref}
\usepackage{enumitem}
\setlist[enumerate]{label*=\arabic*.}
\usepackage{csquotes}
\usepackage{complexity}
\usepackage{xcolor}
\usepackage{subcaption} 
\usepackage{caption}
\usepackage{bbm}
\usepackage{amsbsy}
\usepackage{aligned-overset}
\usepackage{tikz}

\fancypagestyle{firststyle}
{
   \fancyhf{}
   \rhead{\includegraphics[height=1.4cm]{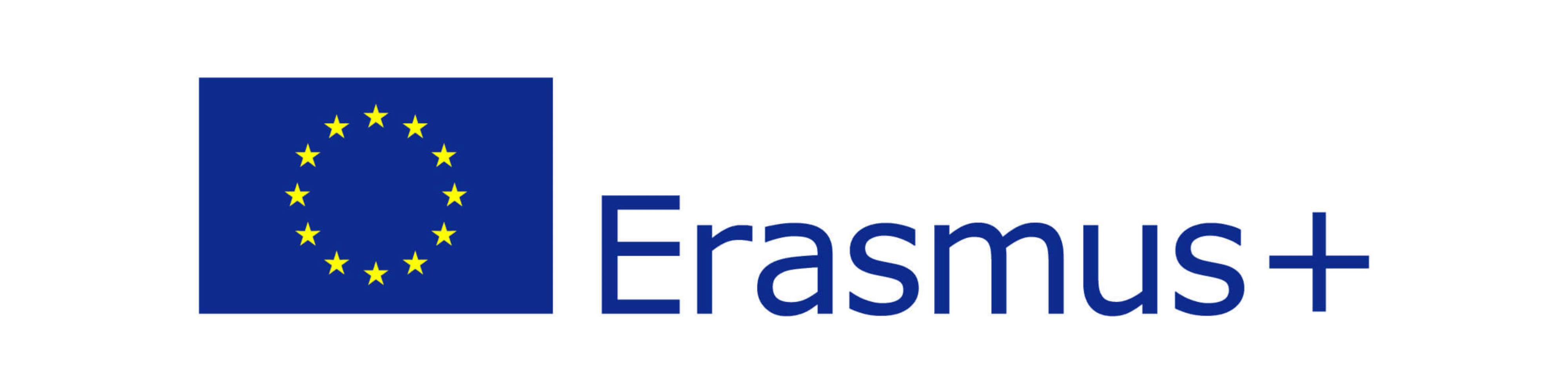}}
   \lhead{\includegraphics[height=1.4cm]{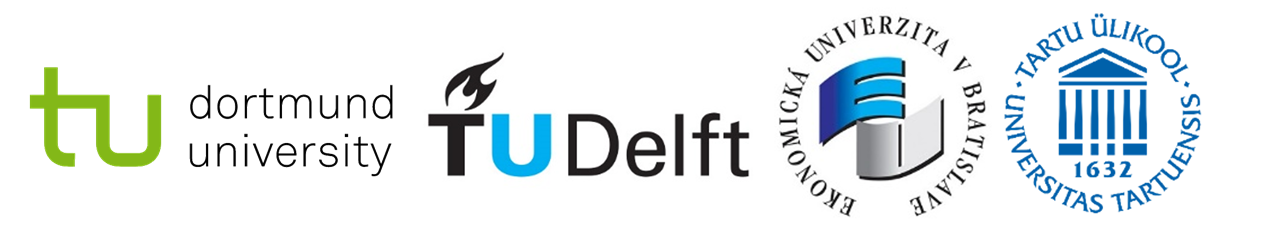}}
   \renewcommand{\footrulewidth}{1pt}
   \cfoot{\thepage}
}

\usepackage{hyperref}
\usepackage{lipsum}
\usepackage{afterpage}
\usepackage{multicol}
\usepackage[paper=portrait,pagesize]{typearea}

\begin{document}

\title{\textbf{MachineLearnAthon: An Action-Oriented Machine Learning Didactic Concept}}

\author[1]{Michal Tkáč}
\author[1]{Jakub Sieber}
\author[2]{Lara Kuhlmann}
\author[2]{Matthias Brueggenolte}
\author[2]{Alexandru Rinciog}
\author[2]{Michael Henke}
\author[3]{Artur M. Schweidtmann}
\author[3]{Qinghe Gao}
\author[3]{Maximilian F. Theisen}
\author[4]{Radwa El Shawi}

\affil[1]{\small Faculty of Business Economy, University of Economics in Bratislava}
\affil[2]{\small Faculty for Mechanical Engineering, TU Dortmund University}
\affil[3]{\small Department of Chemical Engineering, Delft University of Technology}
\affil[4]{\small Institute of Computer Science, University of Tartu}

\date{}

\maketitle
\thispagestyle{firststyle}

\pagestyle{fancy}
\lhead{Tkáč et al.}
\rhead{MachineLearnAthon: An Action-Oriented ML Didactic Concept}
\renewcommand{\footrulewidth}{1pt}

\begin{abstract}

Machine Learning (ML) techniques are encountered nowadays across disciplines, from social sciences, through natural sciences to engineering.
The broad application of ML and the accelerated pace of its evolution lead to an increasing need for dedicated teaching concepts aimed at making the application of this technology more reliable and responsible.
However, teaching ML is a daunting task.
Aside from the methodological complexity of ML algorithms, both with respect to theory and implementation, the interdisciplinary and empirical nature of the field need to be taken into consideration.
This paper introduces the MachineLearnAthon format, an innovative didactic concept designed to be inclusive for students of different disciplines with heterogeneous levels of mathematics, programming and domain expertise.
At the heart of the concept lie ML challenges, which make use of industrial data sets to solve real-world problems. 
These cover the entire ML pipeline, promoting data literacy and practical skills, from data preparation, through deployment, to evaluation.

\end{abstract}

\section{Introduction}
\label{sec:introduction}
In an era marked by rapid digitalization across various life domains, developing data literacy and specific Machine Learning (ML) skills is becoming crucial~\cite{McKinseyWork}. 
ML is a \enquote{field of study that gives computers the ability to learn 
without being explicitly programmed}~\cite{mahesh2020machine} and comprises a subset of Artificial Intelligence (AI) techniques. 
In recent years, it has led to many technological advancements, e.g. in the context of Industry 4.0~\cite{lee2021technological}, large language models,~\cite{vaswani2017attention} such as chat GPT, or object detection frameworks, such as YOLO~\cite{redmon2016you}.
These advances span a large area of applicability but come at a risk: Given their data-driven nature, ML methods are inherently subject to a data selection bias~\cite{donovan2018algorithmic}.
ML competencies, particularly with respect to algorithm contextualization and evaluation, are vital for creating not only efficient and practical but also safe and fair ML solutions. 
To address complex societal and industrial challenges, 
 ML teams should be comprised not only of ML but also of application domain experts. 

Educating domain experts in ML can be a demanding task. 
ML spans several paradigms, from supervised, through unsupervised to reinforcement learning, with each category containing a plethora of algorithms of considerable complexity. 
To make matters worse, owing to the data-driven nature of the field, considerable additional skills are required to acquire and manipulate the algorithm inputs, and evaluate the achieved results~\cite{domingos2012few}.
As such, ML primarily involves applied learning, often intertwined with intricate mathematical theory. 


This paper aims to systematically search the existing literature for teaching principles of ML. Based on the findings, we derive an innovative didactic framework, the MachineLearnAthon concept, that expands on traditional ML challenges to better accompany non-ML-domain student bodies.
Unlike standard ML challenges, which are usually aimed at those already skilled in ML and programming, the MachineLearnAthon concept is designed to be inclusive for students with diverse levels of ML, statistics, and programming expertise. 
The inclusivity combined with modern teaching principles and a strong emphasis on active learning makes the MachineLearnAthon concept highly engaging. 
Gamification elements, e.g. leaderboards, and real-world industrial problems serve to enhance student motivation. 
The approach teaches data literacy and practical ML project skills, from data preparation over deployment to evaluation, while also raising awareness about the potential risks associated with ML. This work is part of the Erasmus+ project MachineLearnAthon. 






The structure of this paper is organized as follows: Section~\ref{sec:Literature Review} introduces the details of the literature review on ML teaching concepts, the didactic design is illustrated in Section~\ref{sec:Didactical design}. Finally, we draw our conclusions and highlight next steps in Section~\ref{sec:Conclusion}.

\section{Systematic Review of Machine Learning Teaching Formats}
\label{sec:Literature Review}
To underline the gap in ML teaching methodology for higher learning institutions, we systematically review related concepts.
Using a systematic search strategy we filter relevant articles from the plethora of academic publications indexed by the Scopus search engine. 
This academic publications database was selected because of its comprehensive indexing of publication venues and its reproducible queries~\cite{mongeon2016journal}. 
By setting clear inclusion criteria and employing a specific search syntax, we narrowed down over a thousand articles to the ones most pertinent to our aim. 
The process reveals a limited number of studies that directly address our core interest.
This chapter discusses the steps taken to arrive at these findings and what they suggest about the current state of ML education.

\subsection[Methodology]{Review Methodology}
\label{subsec:alpha}

\begin{figure*}
\begin{center}
\includegraphics[scale = 0.14]{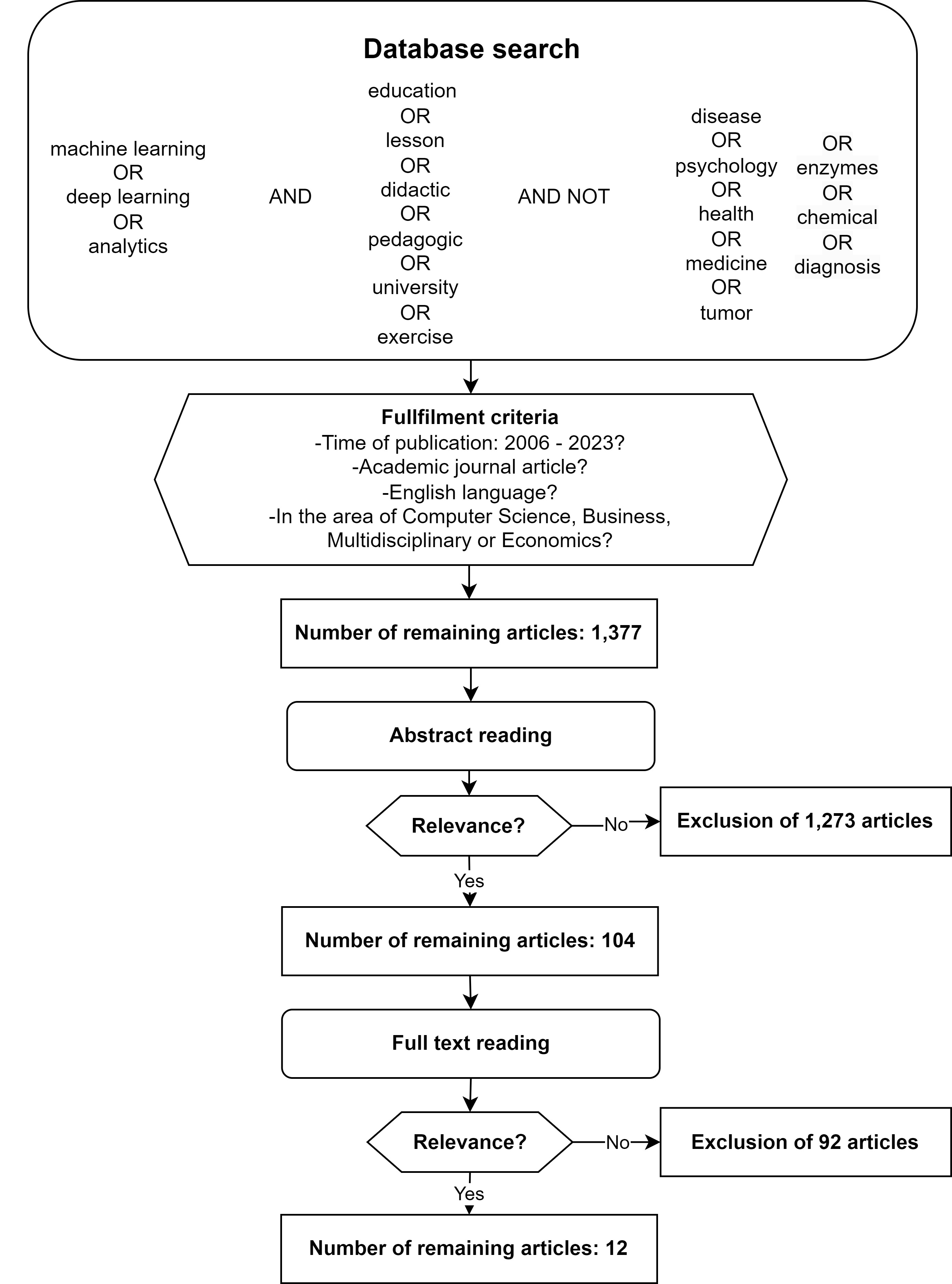}
\caption{Process of literature review}
\label{fig:lit_review}
\end{center}
\end{figure*}

Figure~\ref{fig:lit_review} succinctly depicts our systematic approach. 
We started by creating a search string syntax. 
The string is intentionally designed to be broad such that the risk of missing pertinent articles is minimized. 
To identify and examine publications pertaining to approaches for teaching ML methods, we divide the search string into two segments from the following two semantic fields:
\begin{itemize}
    \item ML segment represented by words: ``machine learning'', ``deep learning'', ``analytics''
    \item Educational segment includes words: ``education'', ``lesson'', ``didactic'', ``pedagogic'', ``university'', ``exercise''
\end{itemize}

We utilized the logical conjunction \enquote{AND} to bridge these two segments, while \enquote{OR} was employed to link the keywords within each individual segment. 
Following the search syntax application, we established stringent inclusion and exclusion parameters. 
These parameters are pivotal in discerning which articles are to be advanced to the next stage of scrutiny. 
The selection criteria were anchored on scholarly, peer-reviewed articles penned in English, with an explicit mention of concrete ML techniques. 
The scope of our research is demarcated to encompass disciplines such as \enquote{Computer Sciences}, \enquote{Business}, \enquote{Multidisciplinary}, and \enquote{Economics}, while omitting any literature accentuating health and medicine-oriented keywords like \enquote{disease}, \enquote{psychology}, \enquote{health}, \enquote{medicine}, \enquote{tumor}, \enquote{enzymes}, \enquote{diagnosis}, 
The temporal boundary for the review was set from the year 2006 onward, marking the dawn of deep learning's emergence.

The exclusion of terms from the medical field was necessary because of a visible bias towards the medical profession recorded by the string not containing the negation.
This bias is most likely due to the inclusion of the \enquote{exercise} term in the educational segment.

The application of our search syntax to the academic databases yields a total of 1,377 articles. 
These articles underwent a preliminary review, 
whereby the abstracts were scrutinized for relevance. 
This rigorous abstract evaluation decreases the number down to 104 articles, which were then subjected to a full-text review. 
This exhaustive review process further distilled the selection to a mere 12 scholarly articles. 
The large number of articles filtered out by the review process can be explained by considering the two possible dependencies between the two conjoined search string segments. 
Most articles discuss the use of machine learning for predicting scholarly outcomes. 
However, the present study's goal is to investigate how pedagogical determine the teaching of ML.
The 12 final articles, which are indexed by Table \ref{tab:Papers lit review}, specifically focus on pedagogical strategies pertinent to the instruction of ML. 
This striking constriction in number from the initial pool starkly highlights a significant gap within the existing literature, signaling an imperative need for an enhanced scholarly focus on didactic approaches to ML education in institutions of higher education.



\subsection[Results]{Results}
\label{subsec:beta}


The review encompasses 12 articles published between 2018 and 2022, each offering unique insights into the methods, applications, and pedagogical strategies employed in the field of ML education. 
From the development of AI education models for non-computer majors to the integration of ML in business analytics and audit curricula, the articles collectively highlight the multifaceted nature of ML teaching. 
They address various learning environments, from classroom-based settings to online platforms, and examine the effectiveness of different teaching methods, including hands-on projects, collaborative learning, and the use of sophisticated programming languages and tools. 
This review not only reflects the current trends in ML education but also sheds some light on practical ML applications, evaluation methods, and impact on students in various academic stages (from school to university). 
As such, the present elaboration can serve as a seed for future investigations of  educational practices in the rapidly evolving domain of ML.

By analyzing the results of the 12 articles featured in our literature review, several noteworthy trends and gaps in the field of ML education become apparent. 
Firstly, there is a striking lack of papers that delve into specific methodologies for teaching ML, indicating a potential area for further research and development.
This gap is significant given the increasing importance of ML in higher education and its potential for career advancement. 
Secondly, 
a predominant number discusses the use of online, ML-based environments for the assessment and evaluation of students. 
This trend highlights the growing reliance on digital platforms to facilitate learning and underscores the need for robust and interactive online educational tools. 
Thirdly, 
several articles emphasize the integration of ML in business analytics education, where students are encouraged to apply ML techniques to solve specific business problems. 
This direct application of ML in a business context mirrors the challenges business students will face in their careers and provides a strong foundation for understanding the potential of ML to transform industries. 
Furthermore, it is notable that only one paper discusses the use of competition or skill comparison as a pedagogical tool. 
This singular mention of competitive learning indicates that this approach is not widely adopted in ML education, despite its potential to enhance student engagement and learning outcomes.

Moreover, the articles describe various learning environments from traditional classroom settings to online platforms, suggesting a flexible approach to ML education that can cater to a diverse range of learning preferences. 
This flexibility is crucial for 
students who may need to balance their education with practical experiences like internships. 
The focus on hands-on projects and collaborative learning highlighted in the articles aligns with the idea of using challenges and competitions in teaching ML. 
By working on tangible projects and competing to develop the best solutions, students can experience firsthand the impact of ML on business decision-making, strategic planning, and operational efficiency. 


\newpage
\KOMAoptions{paper=landscape,pagesize, DIV=18}
\recalctypearea
\newgeometry{left=1cm,right=1cm,top=1.5cm,bottom=1cm} 

\begin{table}[h]
\fontsize{9pt}{9pt}\selectfont
\center
\begin{tabular}{ |m{0.6cm}|m{1.8cm}|m{0.7cm}|m{3.5cm}|m{3.5cm}|m{1.9cm}|m{3cm}|m{7.5cm}| }
\hline
Year&Authors&Use Case&Learning Environment& Teaching Evaluation&Programming Language&Platforms Used&Teaching Process Description\\
\hline
\hline
2018	&Kopcso and Pachamanova \cite{kopcso2017case}&	No&	Undergraduate students, MBA students, and executives&	Survey among students	&R&	Not mentioned&	Suggests ways to frame classroom discussion around the business value of models in data science, predictive analytics, and management science classes
\\
 \hline
 2020	&Marques et al. \cite{marques2020teaching} &	No	&K12 students from primary to high school&	Generally through questionnaires, mostly not systematically evaluated	&Python&	Focus on instructional methods rather than platforms	&Systematic review of ML teaching in schools, analyzing \enquote{Instructional Units} from literature in terms of ML content, teaching methods, and evaluation\\
 \hline
2021	&Alexandre et al. \cite{alexandre2021and}&	Yes&	Citizens 15 years and older, including schools and the public&	Learning analytics, quantitative and qualitative evaluations&	Not mentioned	&MOOC platform	&Discusses an open educational approach to AI using a hybrid MOOC. It focuses on engaging citizens and investigates pedagogical methods and citizen education in AI
\\
 \hline
 2021&Blix et al. \cite{blix2021well}&	No&	Accounting graduates and educators&	Examination of textbooks and online resources	&Not mentioned	&Textbooks and online resources&	Evaluates the integration of data analytics content in prominent auditing textbooks, focusing on technologies, software-based exercises, and alignment with professional standards
\\
 \hline
2021&Brown-Devlin \cite{brown2021teaching}&Yes&	Analytics-focused course for advertising students&	Through course modules and various assignments	&Not mentioned	&Resources, datasets, software&	Provides an overview of teaching an analytics-centered course in a leading advertising program, including descriptions of course modules, assignments, and references to teaching resources and software
\\
 \hline
 2021	&Lee and Cho \cite{lee2021development} &	Yes	&Non-computer majors for general AI education&	Experimenting with AI tools&	Python&	AI education tools, teachable machines&	Discusses classifying ML models and introducing an AI education model using teachable machines for individuals without deep math or computing knowledge\\
 \hline
2021&Lim and Heinrichs \cite{lim2021developing} &	Yes	&Senior-level business students	&Through a marketing analytics project development model and course evaluations	&Not mentioned&	HubSpot's CRM software tools and a learning management system&	Introduces a marketing analytics project development model in a senior-level course. Uses CRM software tools to expose students to data visualizations and analytics\\
 \hline
2021	&Luo \cite{luo2021incorporating}&No&	Auditing educators	&Integration of analytics mindset into the curriculum	&Not mentioned	&Not specified	&Emphasizes the importance of audit data analytics in the audit profession and advocates for auditing educators to integrate an analytics mindset into their curriculum\\
 \hline

2021	&Pudil et al. \cite{pudil2021further}&Yes&	Further education and training of employees&	Analyzing the association between specific educational methods and profitability indicators&	Not mentioned&	Not mentioned&	Focuses on the relationship between specific methods of employee education and financial performance of organizations in the Czech Republic, highlighting the importance of instructing, coaching, mentoring, and talent management\\
 \hline
 2022&Anand et al. \cite{anand2022objectives}&	No&	University students in the business school at the University of Texas, Austin, aged 17-40	&Teaching evaluation, open-ended survey questions, employment outcomes&	Python&	Not mentioned&	Creation of teams, interaction with sponsors, tailoring of in-class learning, execution of business analytics projects, bi-weekly mentoring meetings, project assessments
\\
 \hline
2022&Irgens et al. \cite{irgens2022characterizing}&		No	&Children aged 9-13 at an after-school center	&Pre- and post-drawings	&Scratch, Google Quick Draw&	MIT’s How to Train Your Robot, AI+Ethics for Middle School Curriculum	&Activities included sketching tasks, group algorithm writing, discussions about ML in daily life
\\
 \hline
2022&Kaspersen et al. \cite{kaspersen2022high}&	No&	High school students aged 17-20	&Not clearly specified	&No specific language; tool with GUI&	VotestratesML, an ethics-first learning tool	&Introduction to tools, group work on model creation, discussions on feature selection and algorithm parameters
\\
 \hline

\end{tabular}
\caption{Identified papers in the systematic literature review}
\label{tab:Papers lit review}
\end{table}

\newpage

\KOMAoptions{paper=portrait, pagesize}
\KOMAoptions{DIV=0}
\pagestyle{fancy}



\section{Didactic Design}
\label{sec:Didactical design}

Based on the results of our literature review, we develop a didactic concept for ML, which we present in the following. 
We start by specifying the learning goals. 
Then, we provide some additional theoretical background on  didactic principles. 
Incorporating these principles into the results from our literature review, we present the content and organizational structure of the MachineLearnAthon. 
Finally, we outline how the course can be integrated into university curricula.

The main learning goal of courses following the MachineLearnAthon concept is to teach ML to students with little or no prior programming knowledge. 
This entails the following sub-goals:
\begin{itemize}
  \item Data literacy improvement
  \item Conveying a basic understanding of ML paradigms and widespread models
  \item Developing the skill to employ ML models using Python
  \item Increasing the awareness of risks and limitations associated with ML
  \item Fostering cooperation in working groups (interdisciplinary and international)
  \item Encouraging application-oriented thinking 
\end{itemize}

\subsection{Foundational  Didactic Principles}
\label{subsec:Didactical principles}

As our literature review showed, there has been only little research on how to teach ML competencies to students. 
Due to this research gap, we build on the few findings from the review and on general didactic principles to derive the MachineLearnAthon concept. 
In the following, we elaborate on the pedagical concepts of action orientation, constructivism and problem orientation.

\paragraph{Action Orientation:} When teaching content, a basic distinction can be made in the didactic context between subject-systematic and action-systematic orientation. 
The subject-systematic approach distributes learning objectives and learning content to individual subjects. 
In this way, the learning objects are considered in isolation and treated separately from each other. 
With an action-systematic orientation, learning content can be re-organized on an interdisciplinary basis according to professional action structures and traditional subjects can be dissolved.
The aim is to prepare learners well for professional practice. 
This is achieved through work-related learning situations. 
If this is implemented consistently, it leads to project-like learning in action situations \cite{riedelschelten2000}. 
It is important that the learning process corresponds to a complete action process.
Action-based learning is made up of an interplay of different sub-processes. 
The task-related level consists of the following steps: Clarifying the task or defining the goal, planning, realizing, presenting and evaluating. 
This level requires support from the individual's motivation, organization and intuition. 
In this way, the ability to act in similar situations is built up by meta-cognitive processes transforming concrete experiences into insights \cite{pfäffli2015}. 
Students should also be able to communicate with other students and teachers. 
It is also important that students have the  opportunity to co-operate with other students and act independently \cite{giest2004}.


\paragraph{Constructivism}

Constructivism describes an epistemological position. This means that constructivists are of the opinion that the reality we experience is not a direct reflection of an objectively tangible reality. Instead, it is deeply subjective, as reality is constructed by each individual. The internal construction process is strongly influenced by memories and experiences \cite{siebertreich2005}. Thus, from a learning theory perspective, knowledge cannot be stored and retrieved (cognitivism) or acquired through the reinforcement or attenuation of behaviour (behaviourism). Constructivists believe that knowledge is constructed by the individual \cite{kerres2018}. \\ Prior knowledge plays a very important role here, as new information is linked to experiences and knowledge that have already been incorporated \cite{looiseyal2014}. It follows that the student must become active in order to acquire knowledge. Therefore, constructivism is not only a theory of knowledge but also a theory of action \cite{siebertreich2005}. In addition, constructivism is a learning-centred approach. The student learns in a self-directed manner. The learning environment aims to support students in the development of problem-solving skills. The learning environment must allow students to make decisions regarding learning content, styles and strategies. The teacher supports the students as required and primarily provides the \enquote{tools} for acquiring knowledge \cite{reinmannmandl2006}. Ideally, the individual construction process should not be disturbed \cite{looiseyal2014}. Students should work with authentic problems. In this way, knowledge is acquired directly with application aspects \cite{reinmannmandl2006}.

\paragraph{Problem Orientation}

Problem-orientated learning approaches are used in the design of the activity-based format. 
The focus is on dealing with authentic problems. This corresponds to the findings of constructivism \cite{weber2004} and action-orientation. As a result, students can use problem-based learning approaches to develop skills in dealing with complex problems. The proximity to action-orientation is evident in the structure of problem-based learning approaches. First, students are given a task, e.g. as a problem to be solved. A solution is then developed by analyzing and researching the task.  The solution is then presented and the entire process is analyzed \cite{kerres2018}. Problem-based learning approaches are therefore particularly suitable for developing students' skills in dealing with complex problems \cite{kerres2018}. 

\subsection{Course Structure}
\label{subsec:Course content}

A prominent method emerging within the ML sphere for educational purposes is the incorporation of action-oriented modules, such as open challenges~\cite{Chow2019}. 
Hands-on experience in general, and gamification elements in particular serve to engage the learner interactively.
A well-known example of such a concept is the widespread use of Kaggle~\cite{Kaggle}, a platform where researchers, educators, and companies publish various ML challenges.
Interactive ML education elements, especially ML challenges, are often %
inaccessible to novices in ML and programming owing to the large spectrum of choice, and problem and solution complexity.
Setting up real-world challenges is an intensive task, involving collaboration with companies for data and use case provision, data anonymization and preparation, and detailed use case description. 

To enable the re-use of material while accounting for the significant variance of ML challenges, a micro-lecture format is appropriate. 
This allows the introduction of new, challenge-specific content at a lower cost, since more the general methodological concept does not need to be re-designed.
Using well separated micro-lectures allows for the flexible re-organization of material to tackle novel problems in a cohesive fashion.
Additionally, the micro-lecture format combined with challenges allows educators to more easily adapt to the audience's skill level by varying task difficulty and employing tool introduction units.

Thus, we build our didactic concept on the following assumptions about teaching ML:
\begin{enumerate}
    \item ML is best taught in a \enquote{hands on fashion} using challenges based on real-world problems and data
    \item ML can be operationalized by students of low to intermediate levels of methodological expertise, given a tailored content selection 
    \item Content tailoring (both theoretical -- methods, and practical -- tools) can be best achieved using a micro-lectures format 
    \item Interdisciplinary collaboration should be encouraged so as to bring methodological and domain expertise together
\end{enumerate}
In terms of content, the MachineLearnAthon should encompass the most widespread ML categories across different paradigms along with exemplary models within each.
As such, in terms of supervised learning, the MachineLearnAthon must include Classification and Regression problems. 
Problems from the category of unsupervised learning to be taught should include Association Rule Mining and Clustering. 
These problems along with selected solution algorithms and basic knowledge of ML tools (e.g. Python libraries -- scikit-lean, keras--, or ML modeling programs -- RapidMiner) will empower students to frame and solve many real-world problems.
Additionally, the listed content sets the stage for more advanced techniques from the field of semi-supervized learning, e.g. reinforcement learning, or AutoML.

In term of AutoML, this includes exploring popular frameworks such as AutoSklearn~\cite{feurer2022auto} and TPOT\cite{olson2016tpot}, designed to empower non-experts with ready-made techniques for building effective ML pipelines efficiently. 
Both the theoretical background of these concepts as well as the implementation in Python will be taught. 
The content creation will rely on well-known and often cited books like \enquote{Automated ML: methods, systems, challenges}~\cite{hutter2019automated}, \enquote{Hands-On ML with Scikit-Learn, Keras, and TensorFlow} \cite{geron2022hands} or \enquote{Python ML} \cite{raschka2015python}.

\subsection{Exemplary Course Timeline}
\label{Course organization}

\begin{figure*}[ht]
\begin{center}
\includegraphics[width=\textwidth]{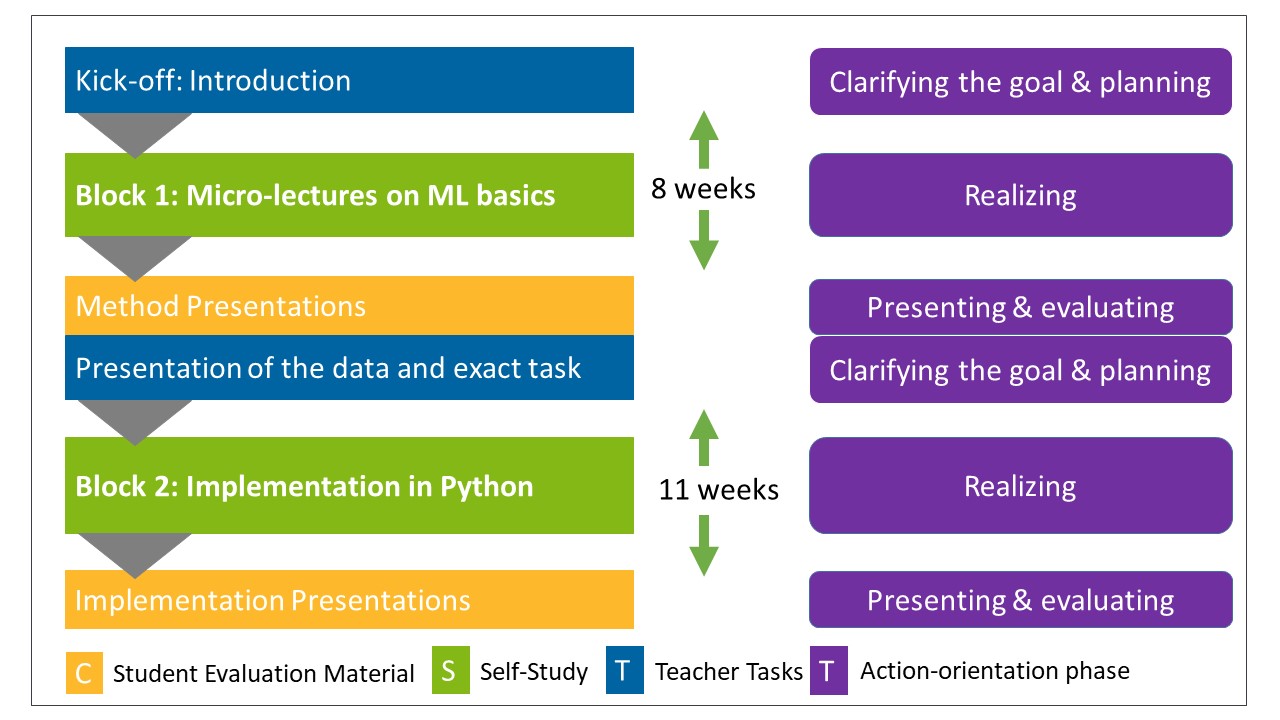}
\caption{Timeline of the ML course}
\label{fig:course_organization}
\end{center}
\end{figure*}

We developed the MachineLearnAthon concept based on the didactic concepts of action- and problem-orientation, and constructivism. The course timeline and organization are displayed in Figure \ref{fig:course_organization}. 
The course consists of two parts, the first is about learning the basics of ML and the second about hands-on application of the recently acquired knowledge. 
Following the action orientation approach, the students will be able to undergo the process of goal clarification, planning, realizing, presenting, and evaluating. 
During the kick-off, the outline and goal of the course will be presented to the students. The goal is to empower the students to solve a real-world use case from the industry with ML techniques on their own. It has been shown that if students internalize learning goals they are more likely to work focused on the task \cite{jiang2011learning}. 

The kick-off will be held on-site. An introduction to the topic will be given and the students will have the opportunity to get to know their teachers and their fellow students. Interaction with their peers was shown to increase their motivation and it facilitates agreements about the group work/ the planning phase \cite{kerres2018}. The groups will consist of three to five participants. This is the recommended group size for action-based learning projects \cite{helle2006project}. 
Moreover, a face-to-face kick-off can help to build a positive relationship between students and teachers, which also increases motivation and participation \cite{ulrich2016gute}. 

In the realization phase of the first part of the course, students are be provided with micro-lectures on relevant topics. Thus, they are able to learn the required methods by tools given by the teacher, as constructivism suggests. In contrast to classical classroom teaching, during which students mainly remain passive, they have to actively deal with the learning material, which helps them to activate their knowledge later when applying it \cite{pfaffli2015lehren}.
The micro-lectures will be provided online. As the literature review showed, online learning proved to be effective and suited for teaching ML. 

The first part of the course will end with a presenting and evaluating phase, in which the different groups present to each other a certain method they learned during the self-study and they will receive feedback from both their peers and teachers. The second part of the course will focus on a hands-on implementation of ML in Python. The concept of problem orientation states the importance of dealing with authentic problems. Subsequently, the students will work on real-world problems and real-world datasets. Again, the part ends with presentations and feedback. The course can be offered at universities as a lecture or laboratory. We suggest to reward the students with 5 ECTS for participation. 



\section{Conclusion}
\label{sec:Conclusion}

This paper presents the innovative didactic MachineLearnAthon concept. 
It addresses the challenges of teaching ML to students with diverse levels of expertise in programming, statistics, and ML. 
The project recognizes the importance of data literacy and specific ML skills in an era dominated by rapid digitization. 
Grounded in a systematic literature review and the  robust didactic principles of action orientation, problem orientation and constructivism, the model emphasizes hands-on learning, interdisciplinary collaboration, and problem-solving skills. 
By incorporating adaptability to diverse skill levels, micro-lectures on ML basics and essential ML tools, the design fosters an inclusive learning environment, bridging the accessibility gap for those new to ML and programming. 
The integration of gamification elements adds a motivational edge, while the commitment to diversity underscores the importance of inclusive participation in the ML domain. 

The work at hand should be regarded as the beginning of a long road towards ML teaching standardization. 
First, an evaluation framework should be conceptualized to allow measuring the success or failure of the pedagogic methodology proposed here.
Secondly, several courses implementing the MachineLearnAthon should be taught at different universities across disciplines. 
To that end real-world datasets and problems need to be acquired from pertinent industry partners. 
Subsequently, the micro-lecture material and tool introduction units need to be developed for the different challenges.
Finally, Using the evaluation framework from the first step, the target audience of the MachineLearnAthon courses as well as the involved teachers the concept can be iteratively revised and improved in subsequent iterations.




\bibliography{arxiv_rl_scheduling_standardization}
\bibliographystyle{ieeetr}

\end{document}